\pdfoutput=1
\documentclass[11pt]{article}

\PassOptionsToPackage{table,xcdraw}{xcolor}
\usepackage[]{EMNLP2023}
\usepackage{tcolorbox}              % for framed, breakable boxes
\usepackage{times}
\usepackage{latexsym}

\usepackage[T1]{fontenc}
\usepackage[utf8]{inputenc}

\usepackage{microtype}

\usepackage{inconsolata}

\usepackage{graphicx} % Required for inserting images
\usepackage{tabularx, makecell, multirow}
\usepackage{booktabs}
\usepackage{geometry}
\usepackage{amsmath, amssymb, amsthm}
\usepackage{float}
\usepackage{caption}
\usepackage{color}

\newcounter{example}[section]
\renewcommand{\theexample}{\nthesection.\arabic{example}}
\newcounter{definition}
\renewcommand{\thedefinition}{\arabic{definition}}
\newenvironment{definition}{
     \refstepcounter{definition}
     {\vspace{1ex} \noindent\bf  Definition  \thedefinition}}{
     \vspace{1ex}} %\hspace*{\fill}\vspace*{1ex}}

\newcounter{theorem}[section]
\renewcommand{\thetheorem}{\nthesection.\arabic{theorem}}
\newcounter{lemma}[section]
\renewcommand{\thelemma}{\nthesection.\arabic{lemma}}
\newcounter{remark}[section]
\renewcommand{\theremark}{\nthesection.\arabic{remark}}
\newcommand{\nthesection}{\arabic{section}}
\newcommand{\beqn}{\begin{eqnarray*}}
\newcommand{\eeqn}{\end{eqnarray*}}

\newcounter{ccc}

\title{Empowering Tabular Data Preparation with Language Models:\\ Why and How?}

\usepackage{mdframed}
\definecolor{lightgray}{gray}{0.95}

\newmdenv[
  backgroundcolor=lightgray,
  linecolor=white,
  innerleftmargin=6pt,
  innerrightmargin=6pt,
  innertopmargin=4pt,
  innerbottommargin=4pt,
  skipabove=10pt,
  skipbelow=10pt,
  roundcorner=0pt,
  linewidth=0pt
]{summarybox}

\author{
\textbf{Mengshi Chen\textsuperscript{1}}, \textbf{Yuxiang Sun\textsuperscript{1}}, \textbf{Tengchao Li\textsuperscript{1}}, \textbf{Jianwei Wang\textsuperscript{2}}, \\
\textbf{Kai Wang\textsuperscript{1}}, \textbf{Xuemin Lin\textsuperscript{1}}, \textbf{Ying Zhang\textsuperscript{3}}, \textbf{Wenjie Zhang\textsuperscript{2}} \\
\textsuperscript{1}Antai College of Economics and Management, Shanghai Jiao Tong University \\
\textsuperscript{2}The University of New South Wales \\
\textsuperscript{3}Zhejiang Gongshang University \\
\texttt{\{chenmengshi,scscsc04,lonelyyy4,w.kai,xuemin.lin\}@sjtu.edu.cn} \\
\texttt{jianwei.wang1@unsw.edu.au, wenjie.zhang@unsw.edu.au, ying.zhang@zjgsu.edu.cn}
}

\begin{document}
\maketitle
\begin{abstract}

% Effective data preparation has become essential across various data-intensive tasks. 
% effective data analysis to support
Data preparation is a critical step in enhancing the usability of tabular data and thus boosts downstream data-driven tasks.
Traditional methods often face challenges in capturing the intricate relationships within tables and adapting to the tasks involved.
Recent advances in Language Models (LMs), especially in Large Language Models (LLMs), offer new opportunities to automate and support tabular data preparation. 
However, why LMs suit tabular data preparation (i.e., how their capabilities match task demands) and how to use them effectively across phases still remain to be systematically explored.
In this survey, we systematically analyze the role of LMs in enhancing tabular data preparation processes, focusing on four core phases: data acquisition, integration, cleaning, and transformation.
For each phase, we present an integrated analysis of how LMs can be combined with other components for different preparation tasks, highlight key advancements, and outline prospective pipelines.
%In conclusion, this survey aims to offer a comprehensive and timely overview of this emerging field, offering insights to guide researchers and practitioners in advancing LM-driven tabular data preparation.

%%{\color{red} How Language Models Act as the Enablers for Tabular Data Preparation: A Survey}

\end{abstract}

\section{Introduction}

Tabular data preparation is the end-to-end procedure of transforming raw, heterogeneous tables into a clean, integrated, and analysis-ready form to enhance data usability. 
Effective tabular data preparation is critical for developing data-driven models, since it prevents the models from falling prey to the garbage in, garbage out (GIGO) syndrome~\cite{dp_book,cite6,cite1,cite2,cite3}. 
Moreover, data preparation demands expert knowledge and involves a substantial labor burden, with data scientists reportedly spending up to 80\% of their time on this task~\cite{hameed2020data}. 
Given these challenges, numerous approaches are proposed to automate and support effective tabular data preparation.

% Moreover, industry surveys consistently report that practitioners spend between 60\% and 80\% of their project time on data preparation tasks, highlighting both its importance and the significant time and effort it demands.

% With the explosive growth of data-driven applications across various domains, effective data preparation has emerged as a cornerstone of modern data-centric tasks.  Garbage In, Garbage Out (GIGO) problem. It typically involves a sequence of stages, including data acquisition, integration, cleaning, and transformation, each addressing specific challenges while building on the previous step to improve overall data quality~\cite{cite4,cite5,cite7}. 
% Usability.

In the early stages, tabular data preparation methods primarily relied on rule-based systems or traditional learning architectures such as tree-based models~\cite{yohannes1999classification, friedman2001greedy} and shallow networks~\cite{kramer1991nonlinear,hawkins2002outlier,batista2003analysis}.
However, these approaches struggle to comprehend the implicit relationships within tables and are difficult to adapt to complex tasks involved in tabular data preparation.
Firstly, they often fail to model fine-grained semantic dependencies, which limits their effectiveness.
Secondly, these methods are typically task-specific and designed with restrictive priors, which restrict their generalization capacity.

\begin{figure*}[t]
  \centering
  \includegraphics[ width=\linewidth]{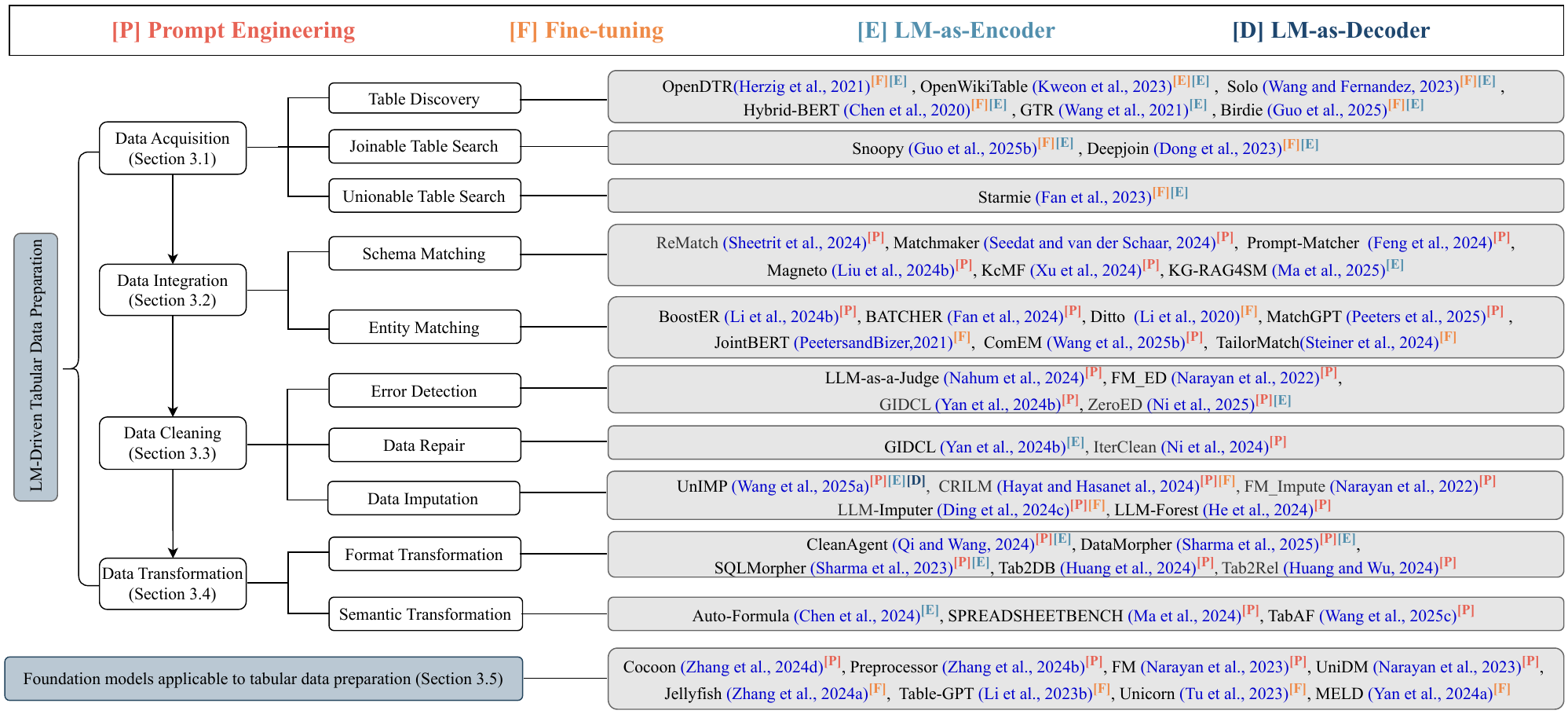}
  \vspace{-6mm}
  \caption{Illustration of the tabular data preparation pipeline, consisting of: data acquisition, data integration, data cleaning and data transformation. Foundation models applicable to data preparation are also explored.}
  \vspace{-6mm}
  \label{fig:overall_framework}
\end{figure*}

Recently, with strong capabilities in semantic understanding, language models (LMs), especially large language models (LLMs) have shown great promise for tabular data preparation~\cite{llmoverview,llmoverview2,liu2023summary}. 
Despite growing interest and notable advancements, existing literature on LM-enabled tabular data preparation often concentrates on isolated tasks or specific technical approaches, thereby lacking a systematic and comprehensive understanding of why LMs are suitable for addressing the inherent complexity of tabular data and how they can be most effectively employed~\cite{lu2025large, DBLP:journals/pvldb/LiZZ24,DBLP:conf/acl/DingQZLLCX0LJ24,DBLP:conf/acl/LongWXZDCW24,zhu2024relational}.
%\textcolor{orange}{Despite this growing interest and the milestones achieved so far, there remains a significant gap in the systematic and comprehensive understanding of how LMs can effectively enable tabular data preparation.}
%Existing literature often focuses on isolated tasks or specific technical approaches, leaving the broader research community without a holistic understanding of how best to employ LMs for tabular data preparation~\cite{lu2025large, DBLP:journals/pvldb/LiZZ24,DBLP:conf/acl/DingQZLLCX0LJ24,DBLP:conf/acl/LongWXZDCW24,zhu2024relational}. 

To address this critical gap, our survey provides a systematic exploration of \textbf{why} and \textbf{how} LMs can empower tabular data preparation. As shown in Figure~\ref{fig:overall_framework}, we structure this survey around four core and closely related tabular data preparation phases: (1) \textbf{data acquisition} that collects data from diverse sources. We focus on \textit{table discovery}, \textit{joinable table search} and \textit{unionable table search}; 
(2) \textbf{data integration} that combines and aligns data. We focus on \textit{schema matching} and \textit{entity matching}; 
(3) \textbf{data cleaning} that handles missing values and errors. We focus on \textit{error detection}, \textit{data repair} and \textit{data imputation};
(4) \textbf{data transformation} that formats and encodes data for analysis. We focus on \textit{format transformation} and \textit{semantic transformation}.

For each of these phases, We first explore the \textbf{``Why''} by analyzing the nature of each task and the strengths of LMs that make them suitable for specific tasks. 
We then dissect the \textbf{``How''} by examining the two main enabling strategies through which these models are employed: (1) LM-centric strategies (e.g., prompt-based and fine-tuning-based) and (2) LM-in-the-loop strategies (e.g., LM-as-encoder and LM-as-decoder), in order to distill generalizable patterns across tasks and frameworks.
While LMs show potential for tabular data preparation, the field is in its early stages. We summarize existing enabling approaches for each phase, pros and cons for different frameworks and highlight promising directions for future research.

\section{Preliminaries}

% \subsection{Problem Definition}

% {\color{red}{See comments}}  \textcolor{blue}{Addressed}

% LLM-driven methods for 

This paper investigates the use of LMs, including LLMs and small language models (SLMs), for tabular data preparation, which transforms raw tables into high-quality, analysis-ready formats.
The detailed introduction of LLM and SLMs is provided in the Appendix~\ref{sec:appendix_lm}.
Tabular data preparation is not a fixed pipeline but is often data-dependent, involving a set of operations to enhance quality.
% and usability.

Let $\mathcal{D}_{\mathrm{lake}}$ denote the initial data lake, containing diverse tables from various sources. A preparation operator set is defined as:
\(
\mathcal{O}_{\mathrm{prep}} = \{O_\text{acq}, O_\text{int}, O_\text{clean},O_\text{trans}, \ldots \}
\) where $O_\text{acq}$ represents data acquisition, $O_\text{int}$ represents data integration, $O_\text{clean}$ is data cleaning, and $O_\text{trans}$ is data transformation. There may be other operators, but these four operators form the core of tabular data preparation and are the main focus of this paper.

% each operator $t \in \mathcal{T}_{\mathrm{prep}}$ can be applied multiple times to different data segments. 

We denote the state of the dataset at time step $k$ as $\mathcal{D}^{(k)}$. 
At the $k$-th step, the system selects an operator $O_k\in\mathcal{O}_{\mathrm{prep}}$ based on current data characteristics, either through expert knowledge~\cite{zhang2003data, mining2006data} or algorithms~\cite{krishnan2019alphaclean,li2021data}. The selected operator is then applied to the $\mathcal{D}^{(k)}$:
% The preparation process proceeds as a sequence of transformation steps:
\[
\mathcal{D}^{(k+1)} \leftarrow \mathcal{M}^{O_k} \left( \mathcal{D}^{(k)} \right),
\quad O_k \in \mathcal{O}_{\mathrm{prep}},
\]
where $\mathcal{M}^{O_k}$ represents an LM-based executor of operator $O_k$. The process continues until the dataset satisfies certain quality criteria or a predefined utility threshold is reached.

\section{Tabular Data Preparation Process}

\subsection{Overview}

\noindent\textbf{Task Overview.} 
 As shown in Figure~\ref{fig:overall_framework}, tabular data preparation is typically comprising data acquisition, integration, cleaning, and transformation. In data acquisition, common tasks include table discovery, joinable table search, and unionable table search. Data integration focuses on schema matching and entity matching to align heterogeneous data sources. Data cleaning addresses error detection, data repair, and imputation to improve data quality. Finally, data transformation covers format transformation and semantic transformation to adapt data for downstream analytics. Definitions for all specific tasks can be found in the Appendix~\ref{sec:appendix_a}.

\begin{figure}[t]
  \centering
  \includegraphics[width=\linewidth]{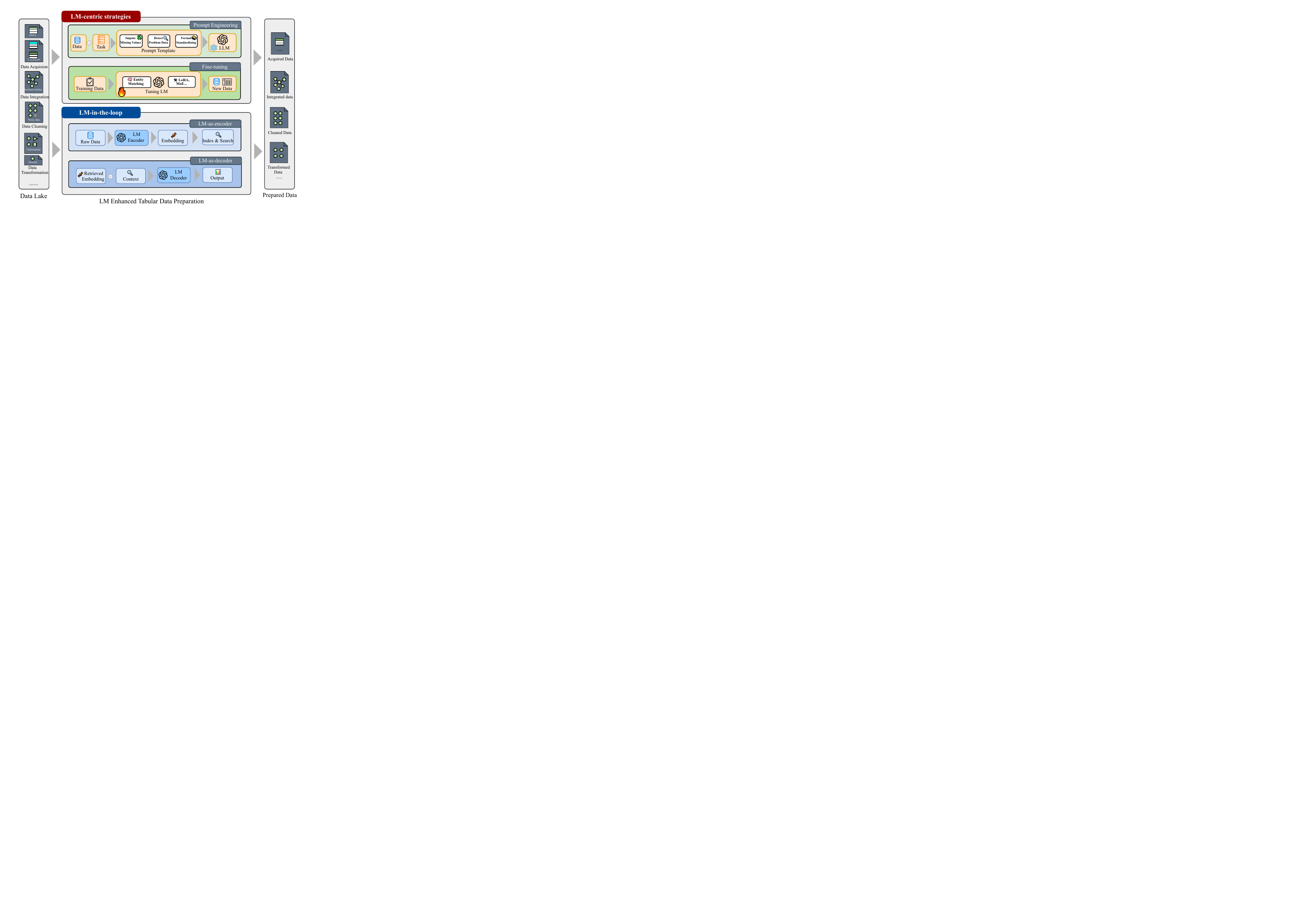}
  \vspace{-5mm}
  \caption{Framework of applying LMs for tabular data preparation: LM-centric (prompt engineering and fine-tuning) and LM-in-the-loop methods (LM-as-encoder and LM-as-decoder).}
  \vspace{-5mm}
  \label{fig:f1}
\end{figure}
\noindent\textbf{Framework Overview.}
As shown in Figure~\ref{fig:f1}, LM-based methods can be grouped into two main strategies: LM-centric strategies and LM-in-the-loop strategies. 
LM-centric strategies use the model directly, either through prompt engineering with curated instructions or by fine-tuning it on task-specific data. In contrast, 
%LM-centric strategies rely primarily on the LM itself, using either prompt engineering or fine-tuning. Prompt engineering guides model behavior through carefully crafted task instructions and examples, leveraging the model’s autoregressive nature without changing its parameters. Fine-tuning instead updates the model with task-specific data to better capture domain patterns.
LM-in-the-loop strategies integrate LLMs or SLMs with other components to combine complementary strengths. LMs usually function as encoders to produce semantic representations or as decoders to generate textual outputs.

Table~\ref{tab:category} presents a taxonomy of LM-enabled tabular data preparation methods.
%They often use LLMs/SLMs as encoders, producing semantic representations, or as decoders, generating outputs from structured signals or intermediate reasoning.

\subsection{Data Acquisition}

\textbf{Touchpoint}. 
Data acquisition is responsible for identifying relevant tables from large data lakes to support subsequent integration, cleaning, and transformation. 
It is often query-driven, where the acquisition process is guided by natural language inputs or query tables. Consequently, state-of-the-art methods typically formulate the problem as a similarity search between the query and data within the data lake, leveraging the semantic understanding capabilities of LMs by incorporating them in an \textbf{LM-as-encoder} paradigm. 
%syx0627
%%\textcolor{orange}{Early methods primarily utilized SLMs for efficient encoding and retrieval, recent approaches build upon this approach by incorporating LLMs to achieve deeper contextual understanding and improved ranking quality.}
% To further improve efficiency, these methods often adopt an indexing strategy and follow a three-stage pipeline: (i) embedding both query and data in the data lake into vector representations, (ii) building indexes over these embeddings offline, and (iii) retrieving the top-$k$ relevant tables through similarity search when an online query is issued.
In this paper, we focus on three core tasks: table discovery, joinable table search and unionable table search.
% In this paper, we focus on three core tasks: table discovery, which identifies relevant tables based on nature language queries, joinable table search, which aims to locate tables that can be joined on shared key columns, and unionable table search, which retrieves tables that can be vertically concatenated based on schema compatibility~\cite{ddiscovery}.
% In this section, we systematically review how LLMs contribute to all three stages of the pipeline, ultimately enhancing the performance of table discovery, joinable search, and unionable search.

\subsubsection{Table Discovery}

\iffalse
\begin{definition}
\textit{(Table Discovery):}
Given a natural language query \( q \) and a table repository \( \mathcal{T} = \{T_1, T_2, \dots, T_N\} \), the goal of table discovery is to retrieve the top-\(k\) tables \( T_t \in \mathcal{T} \) that are most relevant to the query. The relevance is measured by a ranking score \( R_{\text{disc}}(q, T_t) = \mathrm{Rel}(q, T_t) \), where \( \mathrm{Rel} \) captures the semantic alignment between the query and the content (e.g., captions, schemas, and cell values) of the table.
\end{definition}
\fi

% In the context of table discovery, distinguishing between LLM and SLM is critical. LLMs are trained on massive datasets, enabling robust generalization and semantic understanding, whereas SLMs are smaller in scale with fewer parameters and training data, resulting in more limited capabilities. In embedding, indexing, and retrieval, LLMs and SLMs exhibit notable differences: during embedding, LLMs generate more comprehensive semantic representations due to extensive pretraining, while SLMs produce less semantically rich embeddings; in indexing, LLMs leverage large-scale knowledge for semantically meaningful table grouping, whereas SLMs rely on basic indexing techniques; and during retrieval, LLMs use advanced reasoning to rank tables more accurately by understanding query context and intent, while SLMs struggle with ranking precision due to limited semantic comprehension.

% representation-index-search
% select-rerank
Table discovery aims to retrieve the top-$k$ tables from a table repository that are most semantically relevant to a given natural language query.
By employing LM as an encoder, table discovery can be specifically categorized into two approaches according to the embedding strategy: representation-index-search and rerank.

\noindent\textbf{Representation-Index-Search}. This approach typically follows a three-stage pipeline: (i) encoding query and data lake tables separately into embeddings, (ii) constructing indexes over these embeddings in an offline phase, and (iii) retrieving the top-$k$ most relevant tables via similarity search. Variations across existing methods generally lie in the choice of embedding methods and indexing techniques.
OpenDTR~\cite{opendtr} adopts TAPAS~\cite{herzig2020tapas} as the backbone of its encoder, while OpenWikiTable~\cite{wiki} provides several encoder options such as BERT~\cite{bert} and TAPAS. Solo~\cite{solo} introduces a more fine-grained representation by encoding each cell–attribute–cell triplet into a fixed-dimensional embedding.
For indexing, most methods rely on approximate nearest neighbor (ANN) algorithms, such as IVF-PQ ~\cite{ivf} in Solo~\cite{solo}. To further enhance the integration between indexing and retrieval, Birdie~\cite{birdie} proposes a differentiable search index that directly determines.

% Recent works benefit from advances in deep learning by leveraging encoders to represent both queries and tables in a shared vector space, and then top‑\(k\) candidates are retrieved via approximate nearest neighbor search(ANN). Representative systems such as DPR~\cite{dpr}, OpenDTR~\cite{opendtr}, and Solo~\cite{solo} follow this paradigm. However, these methods often suffer from representation bottlenecks and error propagation across modular stages, and they ignore interactions between query and table during encoding.

% To address these limitations, \textbf{Birdie}~\cite{birdie} replaces the traditional three-stage pipeline with a \emph{differentiable search index}. It assigns each table a prefix-aware identifier (\textit{tabid}) using a two-view clustering strategy, and trains an encoder–decoder LLM to directly generate the tabid of the most relevant table from a natural language query—thereby integrating indexing and retrieval into a single sequence-to-sequence model. In addition to retrieval, LLMs are also employed to synthesize diverse, high-quality queries for each table, enhancing training supervision without human annotation. 

\noindent\textbf{Rerank}. In contrast to methods that encode the query and data separately, the rerank approach leverages LLMs/SLMs to jointly encode the query and candidate tables, capturing fine-grained semantic interactions. As it requires computing new embeddings for each query, this approach is generally more time-consuming. To mitigate this, a lightweight retriever first selects a pool of candidate tables, which are then reranked by a joint encoder based on query–table interactions.
Hybrid-BERT~\cite{chen2020table} employs a content selector to extract relevant table segments and uses BERT with a regression layer to predict matching scores. GTR~\cite{wang2021retrieving} introduces a pooling mechanism to integrate the embeddings of the query and candidate table.

%syx0628
%\textcolor{orange}{
\textbf{Pros \& Cons.} Representation-Index-Search approach is highly scalable and efficient but may lose fine-grained alignment information and is less convenient to update when new data is added.
Rerank excels at capturing the correlation between a query and each candidate table by joint encoding, which improves semantic precision. However, it can be computationally intensive and less scalable.
%}

\subsubsection{Joinable Table Search}
\iffalse
\begin{definition}
\textit{(Joinable table search):}
Given a query table \( T_q \) with a specified column \( C_q \), and a table collection \( \mathcal{T} = \{T_1, T_2, \dots, T_N\} \), the goal of joinable table search is to retrieve the top-\(k\) tables \( T_t \in \mathcal{T} \) such that there exists a column \( C_t \in T_t \) maximizing a joinability score: $R_{\text{join}}(C_q, C_t) = \mathrm{JoinSim}(C_q, C_t)$,
where \( \mathrm{JoinSim} \) measures the likelihood that \( C_q \) and \( C_t \) can serve as join keys. 
\end{definition}
\fi
Joinable table search seeks to identify the top-$k$ tables from a collection that contain a column capable of serving as a join key with a query table.

% \noindent\textbf{Lexical relevance}. In the early stage, methods identify joinable tables based on the overlap of column sets, such as Josie~\cite{josie} and LSH Ensemble~\cite{lsh}. They typically compute lexical overlap as a proxy for joinability, without incorporating semantic meaning.

% Rather than computing overlaps between column values, \textbf{semantic embedding-based methods} leverage LLMs to capture the latent semantics of columns. 
% \noindent\textbf{Semantic relevance}. 
Recently, LM-as-encoder is explored to capture semantic information to improve performance. Similar to table discovery, these methods often follow a representation–index–search paradigm. DeepJoin~\cite{deepjoin} fine-tunes SLM-based encoders (e.g., DistilBERT~\cite{disbert}, MPNet~\cite{mpnet}) to capture the latent semantics of columns. It embeds the query column and retrieves joinable candidates from a prebuilt HNSW index~\cite{29}. Snoopy~\cite{guo2025snoopy} also follows this paradigm and further introduces the technique of proxy columns to enhance representations by contrastive learning.
% This approach enables matching even in the absence of direct value overlap, as the model can infer type compatibility and semantic equivalence. 

% \textcolor{orange}{
% \textbf{Pros \& Cons.} Lexical relevance is straightforward and efficient, but its reliance on surface signals makes it less robust when schemas use inconsistent naming or implicit links.
% Semantic relevance complements this by modeling latent relationships and contextual meanings, which improves match quality but comes with higher computational cost and a risk of semantic drift.
% %}
\textbf{Pros \& Cons.} LM-as-encode models latent relationships and contextual meanings, which improves match quality but comes with higher computational cost and a risk of semantic drift.

\subsubsection{Unionable Table Search}

% Unionable table search targets tables with schemas compatible for vertical concatenation, expanding the row space of a query table.
\iffalse
\begin{definition}
\textit{(Unionable Table Search):}
Given a query table \( T_q^U \) and a table collection \( \mathcal{T} \), top-\(k\) table union search retrieves a subset \( \mathcal{T}_q \subset \mathcal{T} \), such that \( |\mathcal{T}_q| = k \), and for any \( T \in \mathcal{T}_q \), \( T' \in \mathcal{T} \setminus \mathcal{T}_q \), it holds that: $R(T_q^U, T) > R(T_q^U, T')$,
where \( R(T_q^U, T_t) \) denotes a table-level unionability score.
% , computed based on the similarity between columns \( C \in T_q^U \) and \( C' \in T_t \). 
\end{definition}
\fi
Unionable table search aims to retrieve the top-$k$ tables from a collection that can be vertically concatenated with a given query table.

% \noindent\textbf{Column-based}. Traditional approaches usually compute unionability scores at the column level, relying on token overlap (e.g., TUS~\cite{tus}) or static embeddings (e.g., D3L~\cite{d3l}). Column-pair similarities are then aggregated to produce table-level scores, without modeling inter-column dependencies.
% However, these models ignore inter-column dependencies and table-wide semantics.

% To overcome the limitations of column-independent matching, \textbf{holistic schema embedding methods} 
% \noindent\textbf{Table-based}. Over r
In recent years, the representation–index–search paradigm has been developed in parallel with the LM-as-encoder approach to capture contextual information across multiple columns.
Starmie~\cite{chai2023starmie} represents tables and reasons about schema-level similarity using an SLM (RoBERTa~\cite{roberta}). It encodes both the query and candidate tables holistically, capturing inter-column relationships. It then employs HNSW~\cite{29} for efficient indexing and introduces bipartite matching to compute table-level similarity scores while enabling pruning during online search.

% Column-based methods are simple and efficient but often ignore inter-column information, which can limit matching accuracy.
%\textcolor{orange}{
\textbf{Pros \& Cons.} 
Table-based methods consider the whole table context and capture richer relationships, but may incur higher computational and memory costs and may overlook fine-grained information.
%}

% \medskip
\begin{summarybox}
\noindent\textbf{Summary and take-away.} 
LLMs/SLMs for data acquisition are typically used as encoders, framing the task as a similarity search. They are often integrated into pipelines following a representation–index–search or re-ranking paradigm. Most existing approaches focus on leveraging SLMs to enhance efficiency.
\end{summarybox}

% Recent advances in unionable table search leverage LLMs to model holistic table semantics, addressing limitations of earlier column-independent matching methods. However, challenges still remain in efficiently scaling to large repositories and handling dynamic schema evolution. 
% Future work could explore lightweight LLM adaptations and incremental learning strategies to bridge this gap.

% is aligned, linked, and consolidated, laying the foundation for reliable data processing
% The process of data integration typically involves two key tasks: schema mapping to align different schemas, and entity linkage to identify records referring to the same entity~\cite{bdi}. 

\subsection{Data Integration}
\textbf{Touchpoint}.
Following data acquisition, data integration focuses on aligning heterogeneous and inconsistent tables to enable subsequent phases.
% While classic algorithms and ML models~\cite{sm_survey2017, survey_classical} fail to handle new inconsistencies or shifts in the data~\cite{fm_wd}, LLMs exhibit robust semantic understanding, strong reasoning capabilities, and the ability to generalize across diverse tasks with minimal task-specific engineering, showing promise for addressing limitations presented above.
Unlike data acquisition, data integration does not involve explicit queries. Instead, every table can be considered a query that needs to be compared with all other tables, resulting in a quadratic complexity.
Hence, state-of-the-art approaches typically \textbf{prompt a LLM} as a reranker or matcher to leverage its deep semantic understanding.
%syx0627
%\textcolor{orange}{
To improve efficiency, recent approaches incorporate SLMs and rule-based heuristics to filter candidats before invoking the LLM.
%}
%in combination with rule-based methods or SLMs that efficiently filter candidates.
% LLMs in data integration are often integrated into the process in a multi-step workflow. State-of-the-art data integration methods leveraging LLMs typically follow a three-stage pipeline:(i) feeding heterogeneous, incomplete, and inconsistent data along with information about the tasks as Prompts or data profiles into LLMs, (ii) generating candidate semantic mappings or decisions through LLMs, often in combination with task-specific frameworks, and (iii) applying and validating the generated semantic correspondences or matching decisions on the data.
In this survey, we focus on surveying schema matching and entity matching.
\subsubsection{Schema Matching}

% Schema mapping is the process of discovering semantic correspondences across heterogeneous data schemas, typically so that data can be integrated or exchanged in a unified manner~\cite{sm_survey2001}. 
\iffalse
\begin{definition}
\textit{(Schema Matching):}
Let $S_s$ and $S_t$ denote the source and target schemas. Each schema $S$ comprises a set of tables $\mathcal{T} = \{T_1, T_2, \ldots, T_m\}$, where each table $T_i$ contains attributes $\mathcal{A}_i = \{A_{i1}, A_{i2}, \ldots, A_{ik}\}$. Schema mapping aims to learn a function $f : \mathcal{A}_s \to \mathcal{A}_t \cup \{\varnothing\}$ that aligns each source attribute from the source schema \( S_s \)  with a target attribute in the target schema \( S_t \) or assigns it to $\varnothing$ if unmatched.

    % Let \( S_s \) and \( S_t \) denote the source and target schemas, respectively. A schema \( S \) consists of a set of tables, represented as \( \mathcal{T} = \{T_1, T_2, \ldots, T_m\} \), where each table \( T_i \) contains a collection of attributes \( \mathcal{A}_i = \{A_{i1}, A_{i2}, \ldots, A_{ik}\} \). The objective of schema mapping is to determine a function \( f : \mathcal{A}_s \to \mathcal{A}_t \cup \{\varnothing\} \) that maps each attribute from the source schema \( S_s \) to a corresponding attribute in the target schema \( S_t \) or assigns it to the empty set \( \varnothing \) when no valid match exists.
\end{definition}
\fi
Schema matching aims to learn a function that aligns attributes from a source schema with corresponding attributes in a target schema.

% Prompting techniques with LLM-based Reranker

\noindent\textbf{Prompt-based LLM reranker.}
% The application of LLMs in schema mapping is predominantly prompt-based, as exemplified by . 
% To address more complex scenarios, recent works rely on 
% (i) \textit{Retrieval-augmented framework}: LLM performance are enhanced by merging semantic retrieval with LLM reasoning. 
% The core idea of these works is to narrow the semantic space by retrieving external knowledge and then conduct semantic alignment using LLMs. 
% \textbf{LLMs-as-encoder}
% In the schema mapping of graph-structured data, 
Related methods usually first generate candidate matches, then design prompts with instructions and candidate matches to further refine them using LLMs~\cite{sm_with_llm_es}. 
They vary in the (1) candidate match generation process and (2) prompt designs.

For the candidate generation, ReMatch~\cite{rematch} converts target schemas into retrievable documents, using semantic similarity to filter potential matches. 
KG-RAG4SM~\cite{kgragsm} leverages external knowledge graphs, encoding the information from graph-structured data and external knowledge graphs for semantic alignment. 
% This enables more accurate semantic matching in the schema mapping of complex data structures.
% (ii) \textit{Hybrid optimization architecture}: Integrating LLMs with external technologies has achieved remarkable results for schema mapping. 
Prompt-Matcher~\cite{caursm} employs traditional schema-matching methods to produce probabilistic candidates. Magneto~\cite{slm-llm} adopts a lightweight SLM for candidate retrieval.
%\textcolor{orange}{
% The use of LLM-generated embeddings in this stage illustrates an approach that can inspire similar techniques for tasks like value normalization and semantic transformation, highlighting how methods designed for one phase can inform solutions in others.
%}

Different prompt designs are also explored to better utilize LLM as a rerank.
Matchmaker~\cite{matchmaker} iteratively refines candidate matches using confidence scores and multi-round prompt feedback. 
KcMF~\cite{kcmf} injects domain knowledge and pseudo-code-like instructions into prompts, explicitly guiding LLMs to reduce confusion and hallucinations. 
% These hybrid approaches collectively overcome the limitations of pure prompt-based methods, adapting LLM-driven schema mapping to diverse and challenging scenarios.

%\textbf{Take Away.}
%Currently, the applications of LLMs in schema mapping are predominantly prompt-based and mainly revolve around retrieval enhancement or hybrid architectures. Future advancements may achieve new progress in other aspects of LLMs.
%\textcolor{orange}{
\textbf{Pros \& Cons.} Prompt-based LLM rerankers leverage small schema sizes and pre-filtering to enable efficient reranking.
They can also use rich contextual cues to resolve ambiguous matches, improving precision. However, poor initial recall of candidates can cause valid matches to be missed.
%}

\subsubsection{Entity Matching}
% Entity linkage, often synonymous with entity resolution or record linkage, refers to the process of identifying and merging records from diverse sources that correspond to the same real-world entity. 
\iffalse
\begin{definition}
\textit{(Entity Matching):}
    Given two sets of records \( \mathcal{R}_s = \{r_1, r_2, \ldots, r_m\} \) and \( \mathcal{R}_t = \{r_1', r_2', \ldots, r_n'\} \), entity matching aims to learn a function \( g: \mathcal{R}_s \times \mathcal{R}_t \to \{0,1\} \), where \( g(r_i, r_j') = 1 \) indicates that \( r_i \) and \( r_j' \) refer to the same entity, while \( g(r_i, r_j') = 0 \) indicates they do not match. 
\end{definition}
\fi
Entity matching identifies whether two records from different sets refer to the same entity.
% Traditional methods include deterministic rules, statistical methods, and machine learning methods~\cite{el}. Currently, given the powerful semantic understanding and logical reasoning abilities of LLMs, an increasing number of studies are exploring LLMs' application to entity linkage tasks, primarily focusing on \textit{prompt engineering} and \textit{fine-tuning method}.
Following the widely adopted block-and-match framework~\cite{wu2023blocker}, LMs are explored as a matcher by prompt engineering and fine-tuning.

\noindent\textbf{Prompt Engineering.}
% Prompt-based methods can effectively tackle entity linkage with minimal domain-specific training.
The method in~\cite{em_using_llm} evaluates various prompts for entity matching and shows that even a simple zero-shot prompt performs well.
% match or exceed classical fine-tuned baselines, revealing substantial potential of  and in-context learning. 
ComEM~\cite{m_c_sr} expands this paradigm by comparing multiple prompting styles: pairwise matching, side-by-side comparison, and selecting one correct record from a set. 
BATCHER~\cite{ce_icl_em} improves cost-effectiveness by batching entity pairs and selecting in-context examples, significantly reducing token usage with little accuracy loss.
BoostER~\cite{booster} adopts selective verification, generating candidate matches with probabilities and refining them using Bayesian adjustment.

\noindent\textbf{Fine-tuning.}
% Earlier studies had already explored fine-tuning pre-trained language models specifically for entity linkage, achieving notable performance improvements.
Fine-tuning is widely used for entity matching. In the early stage, most methods are primarily based on fine-tuning SLMs.
JointBERT\cite{JointBERT} fine-tunes BERT simultaneously on binary matching decisions and multi-class entity classification tasks, leveraging richer supervision signals. Similarly, Ditto\cite{Ditto} demonstrates further gains by enhancing pre-trained transformer models (e.g., BERT, RoBERTa) with targeted textual augmentations and domain-specific knowledge injection. 
% These foundational methods have laid important groundwork for more recent works that fine-tune LLMs for entity linkage.
Method in ~\cite{ft_llm_em} finetune both SLM and LLM and show that properly fine-tuned smaller models consistently improve matching performance, whereas results for LLMs are more variable.
% , suggesting that model scale, domain coverage, and the nature of the fine-tuning data can all impact final accuracy. 

%\textcolor{orange}{
\textbf{Pros \& Cons.} Prompt engineering enables the entity matching models to adapt quickly across different domains, without additional training. However, matching accuracy can be unstable when prompts fail to capture fine-grained differences between ambiguous entities. Fine-tuning improves alignment consistency and task-specific accuracy, but requires labeled pairs and retraining.
%}

\begin{summarybox}
\textbf{Summary and take-away.}
Leveraging the semantic capacity of LMs, both prompting and fine-tuning methods have been explored for data integration. To address efficiency concerns, LLM is often used as a reranker in the retrieval–rerank pipeline or a matcher in the blocker–matcher pipeline.
% Current research on using LLMs for schema mapping and entity linkage mainly relies on existing techniques like prompt-based strategies and fine-tuning, yet lacks comprehensive exploration in optimizing these methods across different data scenarios, and there is a shortage of unified and efficient frameworks that can adapt to diverse data characteristics and complex real-world requirements.
\end{summarybox}

\subsection{Data Cleaning}
\noindent\textbf{Touchpoint}.
Data cleaning aims to identify (error detection) and rectify (data repair and data imputation~\cite{wang2024missing}) errors or inconsistencies in a dataset to improve data quality.
Unlike query-driven data acquisition and pairwise-oriented data integration, data cleaning is typically performed at the instance level.
This prior offers increased flexibility and minimal structural constraints, allowing various strategies to be contextually incorporated into the cleaning process.
%syx0627
%\textcolor{orange}{
Early methods based on SLMs offered efficient but limited support while recent advances leverage LLMs to enable more powerful detection, correction and imputation.
%}
% In this process, LLM orchestration primarily relies on two strategies: \textbf{Prompt-based} and \textbf{LLM-in-the-loop}. 
% The Prompt-based approach designs diverse prompts to harness LLMs' semantic capabilities, enabling them to complete data cleaning tasks by leveraging pre-trained knowledge. 
% The LLM-in-the-loop approach integrates LLMs as tool components for data generation and verification, working iteratively or collaboratively with other techniques. 
% This allows LLMs to adapt to various scenarios, refine results based on feedback, and improve the overall effectiveness of data cleaning by capturing data distributions and utilizing contextual dependencies.
% In this survey, we focus on error detection, data repair and data imputation\cite{zhu2024relational}.

\subsubsection{Error Detection}
% Definition issue 
% Submodular Problems
% Development Trace
% Takeaway for each section
% Error detection is crucial for ensuring data accuracy and quality in downstream tasks.
\iffalse
\begin{definition}
\textit{(Error Detection):}
Error detection aims to identify \(\mathcal{E}=\left\{t_{ij} \in T \mid t_{ij} \text { is erroneous}\right\}\) given the raw dataset ${T}$, where the determination of an “erroneous” cell can be based on various criteria such as violation of data integrity constraints, semantic inconsistencies, or statistical anomalies.
\end{definition}
\fi
Error detection aims to identify erroneous cells on the table based on criteria such as integrity constraint violations and semantic inconsistencies.
% Prompt-based, ZeroED
% Traditional error detection methods include statistical approaches, clustering techniques, outlier detection methods, and constraint-based methods, where constructing a predictive model typically demands a substantial amount of labeled data for model training. However, the zero-shot and few-shot learning capabilities of LLMs have revolutionized this approach. 
Error detection is a discriminative task. LM-based methods typically adopt two strategies: (1) using prompts to reframe the task as generative to leverage LLM knowledge, and (2) using LM-as-encoder to extract features for classifier training.

\noindent\textbf{Prompt-based}. Prompt-based methods have various implementation strategies. 
Some methods directly prompt the LLM to determine whether a given instance is erroneous~\cite{fm_wd}. 
To accelerate error detection, ZeroED~\cite{ni2025zeroed} prompts the LLM to generate rules, which can be efficiently and scalably applied across large datasets.
Furthermore, GIDCL~\cite{gidcl} adopts an iterative approach that leverages LLMs to generate rules for corrupting clean data and synthesizing dirty samples. 
The LLM-as-a-Judge framework~\cite{llm_led} employs an ensemble of LLMs to detect errors in labels in the datasets.

\noindent\textbf{LM-as-encoder}.
In the early stage, SLMs are used to encode data features and train detection models (e.g., Tabreformer~\cite{nashaat2021tabreformer}), balancing efficiency and accuracy. Recently, LLMs are explored. ZeroED~\cite{ni2025zeroed} proposes LLM-based error reason-aware features, which are then used to train a binary classifier.

% Recent works anomaly detection using LLMs are mainly \textbf{Prompt-Based}~\cite{llm_ad_survey}. In these methods, specific prompts are carefully designed and strategically deployed to orchestrate the operation of LLMs, effectively leveraging their semantic understanding and reasoning capabilities.

% SIGLLM\cite{sigllm} uses \textit{role-play prompts} to identify anomalies in time-series data by directing the LLM to look for unexpected patterns. Similarly, LLMAD\cite{llmad} employs contextual learning with \textit{chain-of-thought (CoT) prompts}, which enhances the model’s ability to detect anomalies by providing examples of anomalous behavior within the context of the data. LogPrompt\cite{logprompt} extends this concept to log data, using \textit{CoT reasoning} to spot unusual events or errors in logs. These methods rely on LLMs' pre-trained knowledge and do not require retraining, making them well-suited for applications requiring real-time anomaly detection.
%\textcolor{orange}{
\textbf{Pros \& Cons.} Prompt-based methods enable flexible rule induction and can handle diverse error types without task-specific training, but results may be inconsistent and sensitive to prompt formulation.
LM-as-encoder approaches provide stable, reusable representations for large-scale detection, yet may struggle to capture rare or subtle errors.
%}

\subsubsection{Data Repair}

\iffalse
\begin{definition}
\textit{(Data Repair):}
Given a raw dataset ${T}$ and a set of detected erroneous cells $\mathcal{E}$ identified through the error detection, data repair aims to construct a new dataset ${T'}$ such that all cells $t_{ij} \in \mathcal{E}$ are rectified and $T'$ contains no erroneous values. 
% ${X'}$ is obtained by updating $m_{ij}$ in ${X}$ to $m_{ij}'$.
\end{definition}
\fi
Data repair seeks to rectify detected erroneous cells in a raw dataset to construct a new dataset free from such errors.
Different from error detection, which is discriminative, data repair is essentially a generative task and thus well suited to LMs. 
% LLM-centric methods are explored.
% Traditional data repair methods typically rely on pre-defined rules, constraint-based systems, or manual feature engineering to impute missing or erroneous values. Although these methods can be interpretable, they require extensive domain expertise and often fail when faced with complex or undefined error patterns. 

\noindent\textbf{Prompt-based}.
% The method in~\cite{fm_wd} directly prompts the LLM with few-shot examples to generate repaired data. Further, 
IterClean~\cite{iterclean} enhances this by providing task instructions along with detailed explanations for each identified error to guide the generation of candidate corrections.
% as a \textbf{generator} in the “generate-validate-refine” loop, where an LLM generates candidate corrections that are validated by an error detector, with uncertain cases fed back to the model for adaptive adjustment. 

% Recent approaches integrating LLMs employ the \textbf{LLM-In-The-Loop} orchestration strategy, where LLMs interact with other components in a cyclic or coordinated manner. 
\noindent\textbf{Fine-tuning}.
GIDCL~\cite{gidcl} introduces a new pipeline: an implicit cleaning stage using a fine-tuned local LLM as the corrector, and an explicit cleaning stage that generates correction patterns. A ranking method is then employed to select the most suitable correction.
% In the case of GIDCL\cite{gidcl}, LLMs act as an \textbf{encoder} within this loop. They leverage few-shot learning to infer repair patterns (such as regex rules) from labeled examples, effectively encoding semantic information about the repair task. These inferred patterns are then used in tandem with graph neural networks (GNNs), which capture structural dependencies via hypergraph modeling. The LLMs' encoded semantic knowledge aligns with the structural understanding provided by GNNs, enabling accurate repair of mixed-type data with complex inter-column relationships.

% This approach leverages LLMs’ ability to refine outputs based on contextual feedback, minimizing hallucination and improving consistency, especially in scenarios with sparse labels or ambiguous errors where static methods struggle to generalize. 

% {\color{red}A one-sentence summary. A one-sentence statement of the gap and a promising direction to imporve it}

%\textcolor{orange}{
\textbf{Pros \& Cons.} 
In data repair, prompt-based methods work well for generating simple repair rules and correcting small-scale value errors without extensive training data, but may yield inconsistent results for larger datasets that need consistent corrections. Fine-tuning learns stable repair patterns and ensures consistency across similar errors, but requires curated data and retraining.
%}

\subsubsection{Data Imputation}
\iffalse
\begin{definition}
\textit{(Data Imputation):}
Data imputation aims to impute the unobserved elements in the raw tabular dataset ${T}$, i.e., $m_{ij}=0$ in the mask matrix ${M}$, and make the imputed dataset as close to the real complete dataset (if it exists) as possible. 
\end{definition}
\fi
Data imputation focuses on inferring and filling in unobserved or missing elements.
% Traditional data imputation began with statistical methods like mean/median imputation and multiple imputation (MI). Subsequently, ML methods such as k-NN, matrix factorization, and random forest were developed\cite{bm_di}. These methods require domain knowledge and manual feature engineering, and have limitations in capturing complex data relationships and handling different missingness patterns.
Similar to the data repair, data imputation is also a generative task, and both LLM-centric methods and LM-in-the-loop methods are explored.

% Recent studies have leveraged LLMs for data imputation, with several innovative approaches enhancing accuracy and efficiency, which can be divided into two main categories: those relying on \textit{semantic and contextual reasoning}, and those integrating \textit{graph-based approaches}.

% In addition to prompting-based approaches, 

\noindent\textbf{LM-centric}. 
The method proposed in~\cite{fm_wd} directly prompts the LLM with few-shot examples to perform imputation. Rather than relying on a single prompt and model, LLM-Forest~\cite{di_llm_frst} constructs a bipartite graph for each feature and integrates information from multiple LLM-based few-shot learning "trees" into a forest by merging these graphs. It then performs weighted voting to derive the final imputation result. Fine-tuning methods are also explored. The approach in~\cite{di_llm_rs} applies LoRA to fine-tune the LLM, incorporates existing data as relevant knowledge into the prompt, and subsequently generates candidate values for imputation. Similarly, CRILM~\cite{di_claim} fine-tunes an SLM using the outputs of an LLM, aiming to balance efficiency and accuracy.
% which converts tabular datasets into natural language context formats, uses LLMs to generate context-dependent natural language description words for missing values, thereby creating a context dataset aware of missing values.

% Graph algorithms combined with LLMs has also shown promising prospect when handling data imputation. 

\noindent\textbf{LM-in-the-loop.} 
Beyond relying solely on LLMs, LLM-in-the-loop methods have also been introduced. UnIMP~\cite{di_hmp} first constructs cell-oriented hypergraphs by using the LLM as an encoder to generate feature representations. It then applies high-order message passing to aggregate and propagate information across the dataset, employing the LLM as a decoder to generate imputations based on the aggregated features.
% This approach is designed to handle mixed-type data, including numerical, categorical, and text data, effectively improving imputation results across different data types.

%\textcolor{orange}{
\textbf{Pros \& Cons.} LM-centric approaches can directly infer missing values but may overfit patterns or hallucinate plausible yet incorrect imputations.
LLM-in-the-loop setups combine LLM reasoning with external rules or models, improving reliability and interpretability, though this adds orchestration overhead and requires careful design.
%}

\begin{summarybox}
\textbf{Summary and take-away.}
LMs, as generative models, are particularly well suited for data cleaning. This area has attracted substantial research attention and is progressing rapidly, with both LLM-centric and LM-in-the-loop approaches being actively explored.
% However, current approaches remain fragmented across data types and deployment scenarios. 
% A promising direction for future work is to develop more unified and adaptable LLM-based frameworks that can generalize across diverse real-world settings and improve overall data cleaning effectiveness.
\end{summarybox}

\subsection{Data Transformation}

% {\color{red}highlight the high-level LLM orchestration here, rather LLM capacity. LLM capacity has been highlighted for many times. more technical details are preferred in the main text. similar to Section 3.1}
% Data transformation plays a pivotal role in modern data preparation pipelines, acting as the bridge between raw, often heterogeneous data sources and analysis-ready datasets. Data transformation can be broadly divided into two complementary sub-tasks: format transformation, which focuses on reshaping or restructuring the data, content transformation which addresses the alignment of values and semantics within specific columns. As data sources and formats grow more complex, researchers~\cite{applications_and_challenges} have explored the robust semantic understanding and reasoning capabilities of LLMs to greatly reduce human effort in designing, customizing, and maintaining transformation pipelines. State-of-the-art data transformation methods leveraging LLMs typically follow a three-stage pipeline:
% (i) feeding raw data with its schema and transformation goals as \textbf{Prompts} or data profiles into LLMs,
% (ii) having \textbf{LLM-as-encoder} generate transformation rules or intermediate representations based on the input,
% and (iii) applying and validating the generated transformations on the data.
\noindent\textbf{Touchpoint}. In contrast to data-centric phases like data acquisition, data integration, and data cleaning, data transformation is more task-oriented and instruction-driven, aiming to bridge the gap between raw data and downstream tasks.
%syx0627
%\textcolor{orange}{
While earlier studies leveraged SLMs for encoding-based transformation, recent advances increasingly turn to LLMs for their stronger instruction-following abilities which make 
%}
%In recent years, LLMs, empowered by their autoregressive architecture,have demonstrated strong capabilities in understanding and executing instructions, particularly those expressed in natural language. This makes 
them well-suited for instruction-driven data transformation, where methods such as prompt engineering and LLM-as-encoder are leveraged to align data with specific requirements. In this paper, we survey formation transformation and semantic transformation.
% Data transformation acts as the bridge between raw, heterogeneous data sources and analysis-ready datasets.
% In this section, it is divided into two sub-tasks: \textbf{Format transformation} focuses on reshaping or restructuring the data, while \textbf{Semantic transformation} focuses on extracting rules, functions, or structured knowledge from the table. 
% State-of-the-art data transformation with LLMs typically follow a three-stage pipeline: 
% (i) feeding raw data, and transformation goals as \textbf{prompts} or data profiles, 
% (ii) letting the \textbf{LLM-as-encoder} generate transformation rules or intermediate forms, 
% and (iii) applying and validating the results on the data.

\subsubsection{Format Transformation}

% Format transformation focuses on converting a given dataset into a new format that is consistent, structured, and ready for downstream analytics. This includes both structural operations and semantic adjustments.
\iffalse
\begin{definition}
\textit{(Format Transformation):}
Given a dataset $D$ with records $\mathcal{R} = \{r_1, r_2, \dots, r_n\}$ and a target format schema $F$, format transformation learns a function $\phi: \mathcal{R} \rightarrow \mathcal{R}^\star$ that restructures each $r_i$ into $r_i^\star$ to conform to $F$.
% Given a dataset $D$ consisting of records $\mathcal{R} = \{r_1, r_2, \dots, r_n\}$ and a target format specification $F$, format transformation aims to learn a function $\phi: \mathcal{R} \rightarrow \mathcal{R}^\star$ such that each record $r_i$ is converted into $r_i^\star$ in accordance with $F$, through structural restructuring and/or attribute-level adjustments.
\end{definition}
\fi
Format transformation aims to restructure records to conform to a specified target format schema.

% Format transformation focuses on aligning the structural layout of datasets, typical operations include pivoting, transposing, and merging or splitting columns so that the resulting data table is consistently formatted and ready for downstream analytics.

% \begin{definition} 
% \textit{(Format Transformation):}
% Given a dataset D consisting of records $\mathcal{R} = \{r_1$, $r_2$, \ldots, $r_n$\} and a target structural schema F, format transformation aims to learn a function $\phi: \mathcal{R} \rightarrow \mathcal{R}^\star$ such that each record $r_i$ is restructured into $r_i^\star$ in accordance with the layout rules specified by F. 
% \end{definition}

% Early work addresses format transformation using carefully designed operators or pipelines. Auto-Tables~\cite{auto-tables}, for instance, defines a domain-specific language of pivot-like operators and systematically searches for multi-step transformations that convert disorganized tables into more standard relational forms, while HAIPipe~\cite{haipipe} combines human-authored (HI-pipelines) and machine-synthesized (AI-pipelines) steps to reconfigure tables. Although effective for routine tasks, these approaches may require manual intervention whenever encountering novel or complex table structures.

% {\color{red}utilize the four strategies as highlighted in introduction}

LLMs are widely employed in a \textbf{Prompt-based} manner to automate format transformation. 
Researchers show that LLMs, like GPT-3.5 and GPT-4, can automatically recognize non-relational patterns and convert them into structured outputs with few-shot prompting~\cite{rela_tab_llm}. 
This idea is further expanded to multi-table scenarios, taking a higher-level overview zoom-in–zoom-out approach, showing a notable performance in the cross-table transformations~\cite{tab_db_llm}.
% Another line of research focuses on \textbf{LLM-as-encoder} format transformation, bypassing heavy operator definitions. 
Instead of directly generating the transformed content, DataMorpher~\cite{datamorpher} introduces a novel paradigm that first generates code and then executes it for data transformation.
% a zero-shot approach that feeds source data profiles into an LLM to generate pivot or merge commands.
SQLMorpher~\cite{dt_llm_es} proposes a new benchmark and applies similar code generation techniques, demonstrating the effectiveness of LLM-based multi-step transformations.

\vspace{-2mm}
\subsubsection{Semantic Transformation}
\iffalse
\begin{definition}
\textit{(Semantic Transformation):}
Given a dataset $D$ with records $\mathcal{R} = \{r_1, r_2, \dots, r_n\}$, semantic transformation aims to learn a function $\psi: \mathcal{R} \rightarrow \mathcal{K}$ that maps tabular content to structured knowledge such as functions, rules, or logic forms.

% Given a dataset $D$ consisting of records $\mathcal{R} = \{r_1, r_2, \dots, r_n\}$, where each record $r_i$ contains one or more attributes $\{f_{i1}, f_{i2}, \dots\}$, semantic transformation aims to learn a function $\psi: \mathcal{R} \rightarrow \mathcal{R}^\star$ such that each $r_i^\star$ embeds structured semantics derived from the tabular content.
\end{definition}
\fi
Semantic transformation seeks to map tabular content into structured knowledge, such as functions, rules, or logical forms.
% Currently, the application approaches of LLMs in semantic transformation are similar to those in format transformation, namely prompt-based methods and LLM-as-encoder methods.
 % guide LLMs through tailored instructions to facilitate semantic transformation. 
Both the LM-as-encoder method and prompt engineering are explored.

\noindent\textbf{LM-as-Encoder}. Auto-Formula~\cite{chen2024auto} adopts a retrieval-based approach, where tables are encoded using SentenceBERT~\cite{reimers2019sentence} to retrieve relevant spreadsheets and formulas. This retrieval augments the table-to-formula generation process by incorporating semantically similar examples.

\noindent\textbf{Prompt Engineering}. A benchmark is introduced in SPREADSHEETBENCH~\cite{st_ssb} that integrates user intents, tabular data, and example outputs, evaluating the performance of LLMs in generating executable code for complex spreadsheet manipulations using real-user prompts. In addition, TabAF~\cite{st_tabaf} employs a dual-prompting strategy to generate both answers and corresponding formulas for table-based question answering, leveraging perplexity-based selection to optimize the final output.
% Harmonia\cite{st_harmonia} utilizes interactive ReAct prompting to automate data harmonization workflows via LLM agents, addressing tasks like schema matching and value standardization.

% Graph-based RAG\cite{st_gbrag} uses LLMs to encode text into knowledge graphs/trees, enabling subgraph retrieval for improved QA reasoning. Similarly, 
% SpreadsheetLLM\cite{st_spreadsheetllm} leverages LLMs to compress spreadsheets via structural anchors and inverted-index translation, enhancing LLM efficiency for large-table tasks.

%\textcolor{orange}{
\textbf{Pros \& Cons.} LM-as-Encoder approaches generate general-purpose representations that help align semantics, but they may miss fine-grained contextual cues.
Prompt engineering provides flexible control over how semantic mappings are specified, yet its effectiveness can suffer if prompts fail to clearly define complex field dependencies or contextual constraints.
%}

\begin{summarybox}
\textbf{Summary and take-away.}
With the integration of advanced prompt engineering and LLM-as-encoder orchestration, LLMs present a novel tool for data transformation.
LLM-based methods are capable of handling diverse data formats and complex user intents, thereby broadening the scope and scalability of instruction-driven transformation workflows.
\end{summarybox}

%syx0628
\subsection{LMs Applicable to Diverse Tasks}
% {\color{red}detailed categories for these models}
%syx0627
%\textcolor{orange}{
Alongside task-specific methods, unified data preprocessors are also being explored.
%}
% using a unified model to handle diverse tasks.

\noindent\textbf{Prompt Engineering.}
A study in~\cite{fm_wd} demonstrates that GPT-3-175B, the largest variant of GPT-3, achieves strong performance for tabular data preparation even under zero or few-shot prompting.
% They highlight how straightforward textual instructions often suffice for the model to align tabular cells, detect anomalies, or reconcile schemas. 
A retrieval-augmented pipeline is introduced in UniDM~\cite{qian2024unidm}.
Cocoon~\cite{dc_llm} extends this insight by mimicking human cleaning and decomposing data cleaning into sub‐tasks.
The framework in~\cite{llm_dp} integrates prompt techniques like zero/few-shot, batch prompting, and feature selection for tabular data preparation.

% All of these approaches remain purely “prompt‐driven”, collectively illustrating the surprising breadth of data workflows that LLMs can handle.

\noindent\textbf{Fine-Tuning-Based Strategies.}
Many work uses fine-tuning to encode knowledge of tabular data and preparation operations into LLMs.
Table-GPT~\cite{tablegpt} introduces table-tuning on 18 synthetic tasks, enabling strong table-centric reasoning, consistently outperforming GPT 3.5.
Jellyfish~\cite{jellyfish} trains a local LLM on a small multi-task corpus using curated instructions and distilled reasoning from a larger model for data preprocessing.
Unicorn~\cite{unicorn} pioneers a unified model for seven data matching tasks (e.g., entity matching, schema matching). 
% Unicorn unifies heterogeneous elements (strings, tuples, KG entities) by serializing them into structured text prompts, then fine-tunes a DeBERTa backbone with a Mixture-of-Experts (MoE) layer to align cross-task semantics. 
% This architecture enables knowledge sharing across tasks, outperforming task-specific models on 15/20 benchmarks while achieving practical zero-shot generalization for unseen matching problems. 
% Unicorn demonstrates how unified models can balance specialization and adaptability, this hybrid architecture directly inspires resource-efficient extensions: 
A SpreadsheetLLM is introduced in~\cite{st_spreadsheetllm}.
MELD~\cite{eme_llm_dp} introduces a Mixture-of-Experts (MoE) method for low-resource data preprocessing with a router network that selects the top-$k$ experts to handle queries.
% Jellyfish~\cite{jellyfish} trains a local LLM on a small multi‐task corpus. By carefully constructing instruction data and distilling reasoning results from a larger open‐source model, Jellyfish achieves performance often rivaling or exceeding GPT‐3.5 on multiple data tasks, yet remains adaptable to new tasks.

%\textcolor{orange}{
% \textbf{Pros \& Cons.} Prompt engineering offers high flexibility, allowing quick adaptation across diverse tasks and domains without extra training cost.
% Fine-tuning leverage large-scale multi-task learning to improve generalization and task coverage, but require significant computational resources and can be costly to maintain for frequent updates.
%}

\vspace{-2mm}
\subsection{Frameworks for Full Pipeline}
%\textcolor{orange}{
Besides focusing on LMs, recent research has begun to explore how to orchestrate the entire pipeline by adopting multi-agent frameworks that allow multiple agents to work together with planning, reasoning, and execution.
%}
%textcolor{orange}{
AutoPrep~\cite{fan2025autoprep} combines a Planner that generates logical workflows, Programmers that produce step-wise code, and an Executor that runs the code to handle question-aware tabular data preparation. Pipeline-Agent~\cite{ge2025texttopipeline} formulates data preparation as an iterative decision-making process, where an agent selects operations, executes them, and adjusts subsequent steps based on intermediate results.
%}
%syx0628 a little strange to suddenly insert this paragraph?
\vspace{-2mm}
\section{Evaluation and Challenges}
%\textcolor{orange}{
% In this section, we review the current practices for evaluating LM-enabled tabular data preparation and discuss the open challenges related to its effectiveness and cost-efficiency.
 %}

\subsection{Evaluation}
%\textcolor{orange}{
Systematic evaluation starts with reliable benchmarks and well-defined metrics.

\noindent\textbf{Benchmarks.} While existing benchmarks for specific tasks are widely used, such as Matchbench~\cite{matchbench}for schema matching, Imputationbench~\cite{impuben} for missing data imputation and Medec~\cite{medec} for error detection, they mostly target individual subtasks rather than covering the entire tabular data preparation pipeline, there is still a lack of unified benchmarks that capture the whole workflows.  

\noindent\textbf{Metrics.}
The evaluation involves two main categories of metrics: (1) evaluation on data preparation accuracy, and (2) evaluation on downstream task performance. The first category measures the quality of prepared data directly against a ground-truth version, using standard scores like Accuracy, F1-score and Root Mean Squared Error (RMSE)~\cite{ce_icl_em, nashaat2021tabreformer, Ditto}. The second category assesses practical utility by measuring the impact on a downstream task, such as comparing a model's performance when trained on the original versus the prepared data~\cite{di_claim, fan2025autoprep, dt_llm_ds}. Detailed definitions of these metrics are in the ~\ref{sec:appendix_b}.
%traditional metrics such as accuracy or F1-score are often insufficient to capture their full effectiveness. Since these methods depend heavily on instruction following and multi-step generation, it is important to additionally assess factors such as instruction interpretability, plan consistency, and downstream task impact. Despite this, the absence of community-agreed evaluation pipelines makes cross-method comparison difficult and slows progress toward standardization.
%}
\vspace{-3mm}
\subsection{Challenges}

\noindent\textbf{Effectiveness and Robustness.} A key challenge of LM-enabled methods is their effectiveness and robustness since LMs can produce unfaithful or logically inconsistent outputs (i.e., hallucinations). To mitigate this, recent works combine LMs with external retrieval or structured knowledge sources. For example, RAG-based pipelines provide factual grounding to reduce hallucinations, KG-RAG4SM~\cite{kgragsm} and KcMF~\cite{kcmf} demonstrate how integrating knowledge graphs can verify or enrich model predictions, improving stability. However, scalable and generalizable solutions for robust, repeatable orchestration remain an open challenge.

\noindent\textbf{Cost and Scalability.} In real-world data pipelines that demand low latency and predictable resource budgets, balancing effectiveness with efficiency has become a practical priority. Recent research addressed this problem through various techniques. One common direction is to leverage traditional methods or SLMs to filter candidates and reduce the load on LLM components, as seen in frameworks like Magneto~\cite{slm-llm}, ReMatch~\cite{rematch}, and Prompt-Matcher~\cite{caursm}. Another promising approach distills LLM knowledge into lighter SLMs, such as in Jellyfish~\cite{jellyfish} and CRILM~\cite{di_claim}, achieving comparable performance with low inference cost.
\section{Future Directions}

\subsection{Resource-Efficient LLM Pipelines}
LLMs inevitably impose high computational and energy costs despite their strong capabilities~\cite{effi}. 
Moreover, the input size of LLMs is inherently limited, making it difficult to encode the whole table within a single inference pass.
At the same time, the volume of real-world data generated across various domains has grown at an unprecedented rate.
% This tension becomes particularly evident when scaling up data preparation tasks on massive enterprise datasets.
While techniques for parameter-efficient fine-tuning and inference acceleration are proposed, the efficiency is still not competitive compared to traditional learning-based methods. Moreover, it remains challenging to balance accuracy and cost in high-throughput settings. 
% Moreover, real-time applications that frequently call upon large models risk incurring escalating inference bills. Future research might therefore focus on intelligent pipelines that strategically route simpler data preparation tasks to specialized small models while reserving LLM calls for more complex tasks. Methods to cache intermediate representations, reduce redundant computations, or distill model capabilities into leaner architectures are equally important. Such resource-efficient strategies are expected to play a critical role in enabling LLM-based data preparation methods to scale in a practical and environmentally sustainable manner.

\vspace{-1mm}
\subsection{Interactions Across Different Phases}
Most existing work treats tabular data preparation phases as independent modules. 
%\textcolor{orange}{
However, in practice, these phases are tightly coupled and can benefit from joint modeling. For example, cleaning and integration strongly influence each other: improved cleaning can reduce noise and inconsistencies, thus facilitating more accurate data integration. Conversely, effective integration can expose redundancies, conflicts, or outliers across sources, which in turn provides valuable context for more informed and targeted cleaning.
%}
%However, in practice, these phases, such as cleaning and integration, are tightly coupled and can benefit from joint modeling. 
% For example, better cleaning improves integration accuracy, while integrated context aids in detecting errors. 
%Although recent pipelines~\cite{chen2023haipipe} address later stages like transformation, they often overlook earlier phases and lack LLM-based coordination. 
%Coordinating the entire process remains challenging due to the diverse nature of subtasks and the variability in user requirements.
%\textcolor{orange}{
Coordinating the entire process remains challenging due to the heterogeneous nature of phases, each of which may require distinct strategies. Moreover, the variability in downstream requirements further complicates the development of a one-size-fits-all solution.

\vspace{-1mm}
\subsection{An Agentic and Automatic Framework}
Existing methods emphasize model and algorithm design but lack mechanisms for automatic deployment in dynamic environments. Recent work explores LLM-based agents executing multi-step workflows with specialized roles~\cite{dt_llm_ds,eme_llm_dp}, yet these systems often depend on static task decomposition and lack adaptive collaboration. This limits their robustness against evolving demands and imperfect inputs.
Furthermore, LLMs are inherently prone to hallucination, posing significant challenges to the development of end-to-end and fully automated agentic systems.

% To move toward more robust orchestration, future research should investigate frameworks where LLM agents dynamically exchange intermediate outputs, incorporate downstream feedback, and iteratively refine their actions. Incorporating structured communication protocols and adaptive role-switching mechanisms—where agents adjust their responsibilities based on data preparation pipeline needs—can improve coordination and end-to-end performance. Such agentic orchestration frameworks will enable more resilient and context-sensitive LLM-driven data preparation systems.

\vspace{-2mm}
\section{Conclusions}
\vspace{-1mm}
LMs are reshaping tabular data preparation with flexible and effective solutions across acquisition, integration, cleaning, and transformation tasks. This survey offers a comprehensive review of why LMs are suitable to empower tabular data preparation and how LMs act as enablers for it, including LM-centric and LM-in-the-loop strategies.
We examine a broad set of techniques, organize recent developments around shared paradigms, and summarize how existing methods are incorporated into practical workflows with future directions. 
% We also outline future directions, including modality support, knowledge use, phase interaction, and the automatic system.

\section*{Limitations}

This survey focuses primarily on tabular data preparation methods. However, in real-world scenarios, many other types of data, such as text, images, or time-series, also require specific preparation strategies. Our analysis does not comprehensively cover these diverse modalities, and thus some conclusions and techniques may not generalize beyond tabular data.

In addition, tabular data preparation is inherently connected to downstream tasks, including analytics, modeling, and decision-making processes. However, this survey explicitly limits its scope to the preparation of data itself, without a detailed discussion of these related downstream applications. The interaction and integration with these tasks remain an important area for future exploration.

% \section*{Ethics Statement}

% \section*{Acknowledgements}

\bibliography{anthology,custom}
\bibliographystyle{acl_natbib}

\newpage
\appendix

\section{Language Models}
\label{sec:appendix_lm}

Language models (LMs), often transformer-based~\cite{DBLP:conf/nips/VaswaniSPUJGKP17}, are trained on large text datasets to understand and generate human language. They operate auto-regressively, predicting the next word based on prior context as follows:

\begin{equation}
\label{equ:autoregressive}
\begin{aligned}
   \theta_{LM} = \mathop{\arg\max}\limits_{\theta}\sum\limits_{i} \log P(d_i|d_{i-1},\cdots;\theta)
\end{aligned}
\end{equation}
Where $d_i$ is a text data token and $\theta_{LM}$ is the parameter of LM.
In the early days, LMs were primarily small language models (SLMs), such as T5~\cite{raffel2020exploring} and Bert~\cite{devlin2019bert}. Recently, driven by the scaling law~\cite{kaplan2020scaling, henighan2020scaling}, which suggests that larger models tend to perform better, LLMs, such as GPT-4~\cite{gpt4}, LLaMA~\cite{llama} and Claude~\cite{claude2023}, are emerging.% in the literature.

\section{Task Definitions}
\label{sec:appendix_a}

\subsection{Data Acquisition}
\begin{definition}
\textit{(Table Discovery):}
Given a natural language query \( q \) and a table repository \( \mathcal{T} = \{T_1, T_2, \dots, T_N\} \), the goal of table discovery is to retrieve the top-\(k\) tables \( T_t \in \mathcal{T} \) that are most relevant to the query. The relevance is measured by a ranking score \( R_{\text{disc}}(q, T_t) = \mathrm{Rel}(q, T_t) \), where \( \mathrm{Rel} \) captures the semantic alignment between the query and the content (e.g., captions, schemas, and cell values) of the table.
\end{definition}

\begin{definition}
\textit{(Joinable table search):}
Given a query table \( T_q \) with a specified column \( C_q \), and a table collection \( \mathcal{T} = \{T_1, T_2, \dots, T_N\} \), the goal of joinable table search is to retrieve the top-\(k\) tables \( T_t \in \mathcal{T} \) such that there exists a column \( C_t \in T_t \) maximizing a joinability score: $R_{\text{join}}(C_q, C_t) = \mathrm{JoinSim}(C_q, C_t)$,
where \( \mathrm{JoinSim} \) measures the likelihood that \( C_q \) and \( C_t \) can serve as join keys.
\end{definition}

\begin{definition}
\textit{(Unionable Table Search):}
Given a query table \( T_q^U \) and a table collection \( \mathcal{T} \), top-\(k\) table union search retrieves a subset \( \mathcal{T}_q \subset \mathcal{T} \), such that \( |\mathcal{T}_q| = k \), and for any \( T \in \mathcal{T}_q \), \( T' \in \mathcal{T} \setminus \mathcal{T}_q \), it holds that: $R(T_q^U, T) > R(T_q^U, T')$,
where \( R(T_q^U, T_t) \) denotes a table-level unionability score.
\end{definition}

\subsection{Data Integration}

\begin{definition}
\textit{(Schema Matching):}
Let $S_s$ and $S_t$ denote the source and target schemas. Each schema $S$ comprises a set of tables $\mathcal{T} = \{T_1, T_2, \ldots, T_m\}$, where each table $T_i$ contains attributes $\mathcal{A}_i = \{A_{i1}, A_{i2}, \ldots, A_{ik}\}$. Schema mapping aims to learn a function $f : \mathcal{A}_s \to \mathcal{A}_t \cup \{\varnothing\}$ that aligns each source attribute from the source schema \( S_s \) with a target attribute in the target schema \( S_t \) or assigns it to $\varnothing$ if unmatched.
\end{definition}

\begin{definition}
\textit{(Entity Matching):}
Given two sets of records \( \mathcal{R}_s = \{r_1, r_2, \ldots, r_m\} \) and \( \mathcal{R}_t = \{r_1', r_2', \ldots, r_n'\} \), entity matching aims to learn a function \( g: \mathcal{R}_s \times \mathcal{R}_t \to \{0,1\} \), where \( g(r_i, r_j') = 1 \) indicates that \( r_i \) and \( r_j' \) refer to the same entity, while \( g(r_i, r_j') = 0 \) indicates they do not match.
\end{definition}

\subsection{Data Cleaning}

\begin{definition}
\textit{(Error Detection):}
Error detection aims to identify \(\mathcal{E}=\left\{t_{ij} \in T \mid t_{ij} \text { is erroneous}\right\}\) given the raw dataset ${T}$, where the determination of an “erroneous” cell can be based on various criteria such as violation of data integrity constraints, semantic inconsistencies, or statistical anomalies.
\end{definition}

\begin{definition}
\textit{(Data Repair):}
Given a raw dataset ${T}$ and a set of detected erroneous cells $\mathcal{E}$ identified through the error detection, data repair aims to construct a new dataset ${T'}$ such that all cells $t_{ij} \in \mathcal{E}$ are rectified and $T'$ contains no erroneous values.
\end{definition}

\begin{definition}
\textit{(Data Imputation):}
Data imputation aims to impute the unobserved elements in the raw tabular dataset ${T}$, i.e., $m_{ij}=0$ in the mask matrix ${M}$, and make the imputed dataset as close to the real complete dataset (if it exists) as possible.
\end{definition}

\subsection{Data Transformation}

\begin{definition}
\textit{(Format Transformation):}
Given a dataset $D$ with records $\mathcal{R} = \{r_1, r_2, \dots, r_n\}$ and a target format schema $F$, format transformation learns a function $\phi: \mathcal{R} \rightarrow \mathcal{R}^\star$ that restructures each $r_i$ into $r_i^\star$ to conform to $F$.
\end{definition}

\begin{definition}
\textit{(Semantic Transformation):}
Given a dataset $D$ with records $\mathcal{R} = \{r_1, r_2, \dots, r_n\}$, semantic transformation aims to learn a function $\psi: \mathcal{R} \rightarrow \mathcal{K}$ that maps tabular content to structured knowledge such as functions, rules, or logic forms.
\end{definition}

\section{Metrics Definitions}
\label{sec:appendix_b}

\begin{definition}
\textit{(Accuracy):}
For a given dataset of \( n \) instances \( \{(x_i, y_i)\}_{i=1}^n \), where \( y_i \) is the true label for input \( x_i \), let \( \hat{y}_i \) be the predicted label. Accuracy is the fraction of predictions that are correct. Using an indicator function \( \mathbb{I}(\cdot) \), it is formally defined as:
$$ \text{Accuracy} = \frac{1}{n} \sum_{i=1}^{n} \mathbb{I}(\hat{y}_i = y_i) $$
\end{definition}

\begin{definition}
\textit{(F1-score):}
In a binary classification context, let TP, FP, and FN be the counts of True Positives, False Positives, and False Negatives. Let:
$$ \text{Precision} = \frac{\text{TP}}{\text{TP} + \text{FP}} \quad , \quad \text{Recall} = \frac{\text{TP}}{\text{TP} + \text{FN}} $$
The F1-score is calculated as:
$$ \text{F1-score} = 2 \cdot \frac{\text{Precision} \cdot \text{Recall}}{\text{Precision} + \text{Recall}} = \frac{2\text{TP}}{2\text{TP} + \text{FP} + \text{FN}} $$
\end{definition}

\begin{definition}
\textit{(Root Mean Squared Error - RMSE):}
For a numerical prediction task, let \( Y = \{y_1, \ldots, y_n\} \) be the set of \( n \) true values and \( \hat{Y} = \{\hat{y}_1, \ldots, \hat{y}_n\} \) be the set of predicted values. RMSE is defined as:
$$ \text{RMSE} = \sqrt{\frac{1}{n} \sum_{i=1}^{n} (y_i - \hat{y}_i)^2} $$
\end{definition}

\begin{definition}
\textit{(Downstream Task Performance):}
Let \( \mathcal{M}_{\text{down}} \) be a downstream model (e.g., a classifier) and \( \text{Metric}_{\text{down}} \) be its evaluation metric (e.g., Accuracy). The performance is determined by training \( \mathcal{M}_{\text{down}} \) on \( \mathcal{D}_{\text{prep}} \) and original raw data \( \mathcal{D}_{\text{ori}} \). The performance of data preparation is the performance gap between prepared data and original data. The performance is computed as:
$$
\begin{aligned}
\text{Performance} =\ & \text{Metric}_{\text{down}}\big( \mathcal{M}_{\text{down}}(\mathcal{D}_{\text{prep}}) \big) \\
& -\ \text{Metric}_{\text{down}}\big( \mathcal{M}_{\text{down}}(\mathcal{D}_{\text{ori}}) \big)
\end{aligned}
$$
\end{definition}

\section{Taxonomy Table}
\label{sec:appendix_c}

A taxonomy table, containing information about data preparation phases, specific tasks in each phase, related models, and relevant framework and highlights is given in Table~\ref{tab:category}.

\begin{table*}[htbp]
\centering
\resizebox{\linewidth}{!}{
\begin{tabular}{c|c|ccc}
\toprule
\textbf{Phase} & \textbf{Task} & \textbf{Model} & \textbf{Framework} & \textbf{Highlight} \\
\cline{1-5}
%%%%%%%%%%%%%%%%%%%%%%%%%%%%%%%%%%%%%%%%%%%%%%%%%%%%%%%%%%%%%%%%%%%%
%%% ------------------ Data Discovery (8 rows) -------------------
\multirow{11}{*}{\textit{Data Discovery}} 
& \multirow{7}{*}{\textit{Table Discovery}} 
   & OpenWikiTable~\cite{wiki} & LM-as-Encoder & Several encoder options (BERT and TAPAS) \\
&  & Solo~\cite{solo} & LM-as-Encoder & Triplet-level cell–attribute embeddings with ANN index \\
&  & Hybrid-BERT~\cite{chen2020table} & \makecell{Fine-Tuning\\\& LM-as-Encoder} & Fine-tuned BERT reranker with content selector \\
&  & GTR~\cite{wang2021retrieving} & LM-as-Encoder & Graph-based joint query–table matching \\
&  & Birdie~\cite{birdie} & \makecell{Fine-Tuning \& LM-as-Encoder\\\& LM-as-Decoder} & Differentiable search index learned end-to-end \\
\cline{2-5}
& \multirow{3}{*}{\textit{Joinable Table Search}} 
   & Snoopy~\cite{guo2025snoopy} & LM-as-Encoder & Proxy-column embeddings via contrastive learning \\
&  & DeepJoin~\cite{deepjoin} & \makecell{Fine-Tuning\\\& LM-as-Encoder} & Fine-tuned column encoder with HNSW retrieval \\
\cline{2-5}
& \textit{Unionable Table Search} 
   & Starmie~\cite{chai2023starmie} & LM-as-Encoder & Table-level embeddings with bipartite matching rerank \\
\cline{1-5}
%%%%%%%%%%%%%%%%%%%%%%%%%%%%%%%%%%%%%%%%%%%%%%%%%%%%%%%%%%%%%%%%%%%%
%%% ------------------ Data Integration (15 rows) -------------------
\multirow{17}{*}{\textit{Data Integration}} 
& \multirow{8}{*}{\textit{Schema Mapping}} 
   & ReMatch~\cite{rematch}  & Prompt Engineering & Retrieval-augmented framework \\
& & Matchmaker~\cite{matchmaker} & Prompt Engineering & Automated self-improving \\
& & Prompt-Matcher~\cite{caursm} & Prompt Engineering & Uncertainty reduction pipeline \\
& & Magneto~\cite{slm-llm} & Prompt Engineering & Combining SLM with LLM \\
& & KcMF~\cite{kcmf} & Prompt Engineering & \makecell{Retrieval-based domain knowledge\\ \& Pseudo-code task instruction} \\
& & KG-RAG4SM~\cite{kgragsm} & \makecell{Fine-Tuning\\\& LM-as-Encoder} & \makecell{Retrieve large external\\ knowledge graph} \\
\cline{2-5}
& \multirow{9}{*}{\textit{Entity Linkage}} 
  & MatchGPT\cite{em_using_llm} & Prompt Engineering & \makecell{Demonstrated the superiority of\\ prompt-based approaches} \\
& & ComEM~\cite{m_c_s} & Prompt Engineering & Hybrid ``match-compare-select'' strategy \\
& & BoostER~\cite{booster} & Prompt Engineering & Uncertainty reduction pipeline \\
& & BATCHER~\cite{ce_icl_em} & Prompt Engineering & Question batching \& Batched prompting \\
& & \cite{ft_llm_em} & Fine-Tuning & Domain-specific fine-tuning \\
& & JointBERT~\cite{JointBERT} & Fine-Tuning & Fine-tuned on different tasks \\
& & Ditto~\cite{Ditto} & Fine-Tuning & \makecell{Targeted textual augmentations\\ \& Domain-specific knowledge injection} \\
\cline{1-5}
%%%%%%%%%%%%%%%%%%%%%%%%%%%%%%%%%%%%%%%%%%%%%%%%%%%%%%%%%%%%%%%%%%%%
%%% ------------------ Data Cleaning (10 rows) ----------------------
\multirow{14}{*}{\textit{Data Cleaning}}
& \multirow{8}{*}{\textit{Data Imputation}}
  & LLM-Imputer\cite{di_llm_rs} & \makecell{Fine-Tuning\\ \& Prompt Engineering} & \makecell{Combining Fine-Tuning\\ and Prompt Engineering} \\
& & CLAIM~\cite{di_claim} & \makecell{Fine-Tuning\\ \& Prompt Engineering} & \makecell{Generating natural language\\ descriptors for missing values} \\
& & LLM-Forest~\cite{di_llm_frst} & \makecell{Prompt Engineering\\\& LM-as-Encoder} & \makecell{Graph-based neighbor\\ identification + LLM forests} \\
& & UnIMP~\cite{di_hmp} & \makecell{Prompt Engineering\\\& LM-as-Encoder} & \makecell{Combining hypergraph message\\ passing with LLM embeddings} \\
\cline{2-5}
& \multirow{4}{*}{\textit{Error Detection}}
  & LLM-as-a-Judge~\cite{llm_led} & Prompt Engineering & Ensembling LLMs to detect label errors \\
& & SIGLLM\cite{sigllm} & Prompt Engineering &  Role-play prompt-driven  \\
& & LLMAD\cite{llmad}  & Prompt Engineering & Chain-of-thought prompt-enhanced \\
& & LogPrompt\cite{logprompt} & Prompt Engineering & Chain-of-thought reasoning \\
\cline{2-5}
& \multirow{2}{*}{\textit{Data Repair}}
  & GIDCL\cite{gidcl}   & Prompt Engineering & \makecell{Few-shot inference of repair patterns\\ with hypergraph modeling} \\
& & IterClean\cite{iterclean} & Prompt Engineering & Iterative generate-validate-refine loop \\
\cline{1-5}
%%%%%%%%%%%%%%%%%%%%%%%%%%%%%%%%%%%%%%%%%%%%%%%%%%%%%%%%%%%%%%%%%%%%
%%% -------------- Data Transformation (6 rows) ---------------------
\multirow{9}{*}{\textit{Data Transformation}}
& \multirow{6}{*}{\textit{Format Transformation}}
  & CleanAgent~\cite{dt_llm_ds} & \makecell{Prompt Engineering\\\& LM-as-Encoder} & \makecell{Automated data standardization via\\ multi-agent LLM collaboration} \\
& & DataMorpher~\cite{datamorpher} & \makecell{Prompt Engineering\\\& LM-as-Encoder} & Automated Python code generation \\
& & SQLMorpher~\cite{dt_llm_es} & \makecell{Prompt Engineering\\\& LM-as-Encoder} & Automated SQL code generation \\
& & Tab2DB~\cite{tab_db_llm} & Prompt Engineering & ``zoom-in''–``zoom-out'' approach \\
\cline{2-5}
& \multirow{2}{*}{\textit{Content Transformation}}
  & DataMorpher~\cite{datamorpher} & Prompt Engineering & Automated Python code generation \\
& & SQLMorpher~\cite{dt_llm_es} & Prompt Engineering & Automated SQL code generation \\
\cline{1-5}

%%%%%%%%%%%%%%%%%%%%%%%%%%%%%%%%%%%%%%%%%%%%%%%%%%%%%%%%%%%%%%%%%%%%
%%% ----------------- General Model (8 rows) ------------------------
\multirow{14}{*}{\textit{General Model}}
& \multirow{14}{*}{\textit{Integrated Tasks}}
  & Cocoon~\cite{dc_llm} & Prompt Engineering & \makecell{Decomposing data cleaning tasks\\ and mimicking human cleaning processes} \\
& & Table-GPT~\cite{tablegpt} & Fine-Tuning & \makecell{First LLM fine-tuned\\ on diverse table tasks} \\
& & LLM Preprocessor~\cite{llm_dp} & Prompt Engineering & \makecell{Integrating a series of SOTA\\ prompt engineering techniques} \\
& & Jellyfish~\cite{jellyfish} &  Prompt Engineering & Knowledge Distillation \\
& & Unicorn~\cite{unicorn} & Fine-Tuning & \makecell{Generic Encoder +\\ Mixture of LLM Experts} \\
& & MELT~\cite{eme_llm_dp} & Fine-Tuning & \makecell{RAG + meta-path search\\ + Mixture of LLM Experts} \\
& & Foundation model\cite{fm_wd} & Prompt Engineering & \makecell{Processing structured data\\ as text and executes tasks} \\

\bottomrule
\end{tabular}
}
\caption{A taxonomy of LM-enabled tabular data preparation methods, grouped by phase and specific task.}
\label{tab:category}
\end{table*}

\end{document}